\documentclass[]{jair}

\usepackage{bbding}
\usepackage{url}
\usepackage{hyperref}
\usepackage{threeparttable} 
\usepackage{multirow}
\usepackage{adjustbox}
\usepackage{makecell}
\usepackage{booktabs, amsmath}

\usepackage{amsfonts}

\usepackage{amssymb}
\usepackage{color}
\usepackage{tikz}
\usepackage[edges]{forest}
\definecolor{hidden-draw}{RGB}{20,68,106}
\definecolor{hidden-pink}{RGB}{255,245,247}

\setcopyright{cc}

\copyrightyear{2025}
\acmDOI{10.1613/jair.1.17625}

\JAIRAE{Huang Xuanjing}
\JAIRTrack{} 
\acmArticle{32}
\acmVolume{83}
\acmMonth{8}
\acmYear{2025}

\RequirePackage[
  datamodel=acmdatamodel,
  style=acmnumeric,
  backend=biber,
  giveninits=true
  ]{biblatex}
\begin{document}

\title{A Survey on Data Selection for LLM Instruction Tuning}

\author{Bolin Zhang}
\authornote{Equal contribution.}
\orcid{0000-0002-6315-1856}
\email{brolin@hit.edu.cn}
\affiliation{%
  \institution{Harbin Institute of Technology}
  \city{Harbin}
  \state{Heilongjiang}
  \country{China}
}

\author{Jiahao Wang}
\orcid{0009-0001-3280-851X}
\authornotemark[1]
\email{jiahaowang0917@gmail.com}
\affiliation{%
  \institution{Institute of Automation, Chinese Academy of Sciences}
  \city{Beijing}
  \state{}
  \country{China}
}

\author{Qianlong Du}
\orcid{0000-0002-7187-2530}
\email{qianlong.du@nlpr.ia.ac.cn}
\affiliation{%
  \institution{Institute of Automation, Chinese Academy of Sciences}
  \city{Beijing}
  \state{}
  \country{China}
}

\author{Jiajun Zhang}
\orcid{0000-0001-5293-7434}
\authornote{Corresponding author.}
\email{jjzhang@nlpr.ia.ac.cn}
\affiliation{%
  \institution{Institute of Automation, Chinese Academy of Sciences}
  \city{Beijing}
  \state{}
  \country{China}
}

\author{Zhiying Tu}
\orcid{0000-0001-8800-4513}
\email{tzy\_hit@hit.edu.cn}
\affiliation{%
  \institution{Harbin Institute of Technology (Weihai)}
  \city{}
  \state{Shandong}
  \country{China}
}

\author{Dianhui Chu}
\orcid{0000-0003-2973-7252}
\email{chudh@hit.edu.cn}
\affiliation{%
  \institution{Harbin Institute of Technology (Weihai)}
  \city{}
  \state{Shandong}
  \country{China}
}

\renewcommand{\shortauthors}{Zhang, Wang, Du, Zhang, Tu \& Chu}

\begin{abstract}
Instruction tuning is a vital step of training large language models (LLMs), so how to enhance the effect of instruction tuning has received increased attention. Existing works indicate that the quality of the dataset is more crucial than the quantity during instruction tuning of LLMs. Therefore, recently a lot of studies focus on exploring the methods of selecting high-quality subset from instruction datasets, aiming to reduce training costs and enhance the instruction-following capabilities of LLMs. This paper presents a comprehensive survey on data selection for LLM instruction tuning. Firstly, we introduce the wildly used instruction datasets. Then, we propose a new taxonomy of the data selection methods and provide a detailed introduction of recent advances, and the evaluation strategies and results of data selection methods are also elaborated in detail. Finally, we emphasize the open challenges and present new frontiers of this task. 
\end{abstract}

\received{14 November 2024}
\received[accepted]{12 April 2025}

\maketitle

\section{Introduction}
Large language models (LLMs), such as PaLM \cite{PaLM}, GPT-4 \cite{gpt4}, and LLaMA \cite{LLaMa}, have demonstrated impressive capabilities across a wide range of natural language understanding and generation tasks. Central to their success is instruction tuning—a process that aligns model outputs with human intent by fine-tuning pretrained models on curated (instruction, response) pairs \cite{LLM_hf, InstructionTuning_for_LLMs}. This technique not only improves the controllability and safety of LLMs but also enables them to adapt to specific domains with minimal computational overhead.

In the existing works \cite{closer, FLAN, FLAN-collection}, instruction tuning mainly focuses on constructing large-scale  instruction-following datasets. These works can be categorized into two types: (1) converting annotated NLP datasets into instruction-style formats using templates (e.g., P3 \cite{P3} and FLAN \cite{FLAN}), and (2) automatically generating instruction-response pairs using powerful models like GPT-3.5 (e.g., Self-Instruct \cite{Self-Instruct}). While these works have

\tikzstyle{my-box}=[
    rectangle,
    draw=hidden-draw,
    rounded corners,
    text opacity=1,
    minimum height=1.5em,
    minimum width=5em,
    inner sep=2pt,
    align=center,
    fill opacity=.5,
    line width=0.8pt,
]
\tikzstyle{leaf}=[my-box, minimum height=1.5em,
    fill=blue!6, text=black, align=left,font=\normalsize,
    inner xsep=2pt,
    inner ysep=4pt,
    line width=0.8pt,
]
\begin{figure*}[!ht]
    \centering
    \resizebox{\textwidth}{!}{
        \begin{forest}
            forked edges,
            for tree={
                grow=east,
                reversed=true,
                anchor=base west,
                parent anchor=east,
                child anchor=west,
                base=left,
                font=\large,
                rectangle,
                draw=hidden-draw,
                rounded corners,
                align=left,
                minimum width=4em,
                edge+={darkgray, line width=1pt},
                s sep=3pt,
                inner xsep=2pt,
                inner ysep=3pt,
                line width=0.8pt,
                ver/.style={rotate=90, child anchor=north, parent anchor=south, anchor=center},
            },
            where level=1{text width=12em,font=\normalsize,}{},
            where level=2{text width=12em,font=\normalsize,}{},
            where level=3{text width=8em,font=\normalsize,}{},
            where level=4{text width=5em,font=\normalsize,}{},
            [
                Data Selection for Instruction Tuning, ver
                [Instruction Sets (\S \ref{instructions})
                    [
                        Alpaca~\cite{alpaca}{, }WizardLM~\cite{WizardLM}{, }Dolly-V2~\cite{DollyV2}{, }P3~\cite{P3}{, }LIMA~\cite{zhou2023lima}{,}
                        Self-Instruct~\cite{Self-Instruct}
                        , leaf, text width=36em    
                    ]
                ]                
                [Data Selection Methods (\S \ref{selectors})
                    [System of indicators (\S \ref{indicators})
                        [
                            INSTRUCTMINING~\cite{InstructionMining}{, } DQ~\cite{DatasetQuantization}{,}\\ InstructionGPT-4~\cite{InstructionGPT-4}
                            , leaf, text width=18em
                        ]
                    ]
                    [Trainable LLMs  (\S \ref{Trainable})
                        [
                            IFD~\cite{IFD}{, }Instruction Backtranslation~\cite{Instruction_Backtranslation}{, }\\Nuggets~\cite{One_Shot_Learning}{, }DIVERSEEVOL~\cite{Self-Evolved}{, }\\TEGIT~\cite{TeGit}{, }ActiveIT~\cite{ActiveIT}
                            , leaf, text width=18em    
                        ]
                    ]
                    [Powerful LLMs  (\S \ref{Powerful})
                        [
                            AlpaGasus~\cite{AlpaGasus}{, }INSTAG~\cite{InsTag}{, } \\ LIFT~\cite{LIFT}{, } DEITA~\cite{DEITA}{, } \\ InstructionNode~\cite{InstructionNode}{, }  WaveCoder~\cite{WaveCoder}
                            , leaf, text width=18em    
                        ]
                    ]
                    [Small Models  (\S \ref{Reward})
                        [
                            MoDS~\cite{MoDS}{, } Coreset-based Selection~\cite{0.5data}
                            , leaf, text width=18em    
                        ]
                    ]
                ]
                [Evaluation Methods (\S \ref{Evaluation})
                    [
                          Wining rate~(\S \ref{Rate}){, }Inner Comparison (\S \ref{Inner_Comparison}){, }External Comparison (\S \ref{External_Comparison}), leaf, text width=36em  
                    ]
                ]
            ]
        ]
        \end{forest}
        }
    \caption{Overview of Data Selection for LLM Instruction Tuning. }
    \label{taxo_of_icl}
\end{figure*}
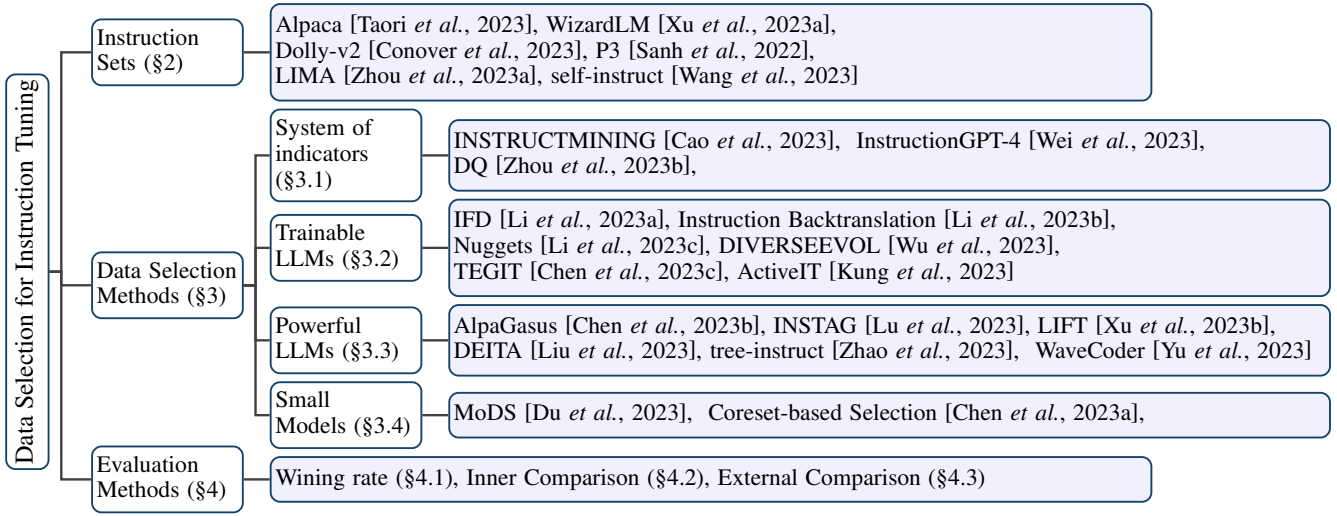

\noindent greatly expanded the availability of instruction data, they often suffer from issues related to data redundancy, limited diversity, and inconsistent quality. Moreover,  these works primarily rely on a large volume of data, which can lead to a significant waste of computational resources and electricity.

In contrast to the previous approaches primarily focused on large-scale data construction \cite{alpaca, WizardLM, DollyV2}, LIMA \cite{zhou2023lima} proposes an approach centered on careful data selection. Their findings demonstrated that a model fine-tuned on merely 1,000 carefully curated instruction examples could achieve performance comparable, and in some cases superior, to models trained on vastly larger, often automatically generated, datasets. This seminal work underscored a critical insight: the quality of instruction data can significantly outweigh its quantity in determining the efficacy of instruction tuning. Consequently, this discovery has prompted a shift in research focus, moving from the large-scale generation of instruction data towards developing better methods for constructing and selecting high-quality, impactful, and efficient datasets.

In these data selection approaches, manual instruction data selection often involves high costs and introduces human bias. As a result, automated methods are crucial for instruction data selection. These methods have shown promising results. For instance, the IFD\cite{IFD} method significantly outperforms the Alpaca model by using only about 5\% of the Alpaca dataset, and it also surpasses the WizardLM model by about 10\%. Using high-quality subsets for fine-tuning not only enhances the instruction-following skills of LLMs but also significantly reduces computational costs and time.

However, the data selection task for LLM instruction tuning is challenging due to the complex factors and multidimensional considerations involved. For example, it is difficult to assess the quality of individual instructions and ensure the overall diversity of the selected data. Another challenge is reducing costs and improving the efficiency of the selection process. In light of these factors, various data selection methods have been developed. Some methods use a system of indicators to evaluate individual data points, while others rely on trainable LLMs or powerful external LLMs. These exploit the LLMs' own capabilities to select instructions. Additionally, methods that use smaller models and design a comprehensive process to achieve a balanced effect across all aspects are noteworthy. 

Despite these promising advances, the field remains fragmented, with no unified understanding of the principles, methods, and trade-offs involved in instruction data selection. In this survey, we aim to fill this gap by providing a systematic overview of instruction data selection for LLMs. We categorize existing methods, analyze their core ideas and assumptions, and highlight open challenges such as the lack of standardized evaluation protocols, inefficiencies in processing large-scale instruction data, and the heavy dependence on powerful foundation models for selection. By synthesizing recent progress and identifying future directions, we hope to establish a foundation for further research and practical development in instruction tuning.

As illustrated by Fig.~\ref{taxo_of_icl}, this paper provides a comprehensive review of existing works on data selection methods for the instruction tuning of LLM. To facilities the community, we maintain a paper list\footnote{\url{https://github.com/Bolin97/awesome-instruction-selector}},  collect commonly instruction sets for data selection. Section \ref{instructions} describes the mainstream datasets with different source and construction methods used in instruction tuning and Section \ref{selectors} describes four types data selection methods in details: indicators sets, trainable LLMs, powerful LLMs and small models. Section \ref{Evaluation} presents the evaluation methods and shows the results of different instruction selection methods. Section \ref{Conclusion} summarizes the paper and emphasize the open challenges and future direction of the instruction selection.

\section{Instruction Datasets}
\label{instructions}

Various instruction tuning datasets (e.g. Self-Instruct and Alpaca), generated by LLMs, offer a wealth of samples without human labor, but their data quality depends on the performance of LLMs and is uncertain. Conversely, the manually curated datasets (e.g. LIMA and Dolly) obtain higher quality through meticulous human selection but are potentially influenced by human biases. Alternative dataset construction methods,  like prompt mapping and evol-instruct, aim to enhance dataset quality and diversity but introduce new challenges in quality assurance. This variability in dataset construction and sourcing significantly affects data quality, highlighting the importance of careful data selection for the instructional tuning of LLM. This section describes the scale and construction procedures of several commonly instruction tuning datasets:

Self-instruct \cite{Self-Instruct} consists of 52,000 training and 252 testing instructions. Initial instructions from seed tasks were selected, categorized, and diversified using InstructGPT \cite{Instructgpt} to generate inputs and outputs through output-first or input-first strategies. Post-processing refined the dataset for uniqueness and relevance, providing a versatile resource for natural language processing applications.

Alpaca \cite{alpaca} is a dataset of 52,002 instruction-response pairs generated using the Self-Instruct framework \cite{Self-Instruct} with the text-davinci-003 model. It is designed to fine-tune LLaMA for improved instruction-following capabilities.

WizardLM \cite{WizardLM} is a dataset containing 250,000 instruction-response pairs generated via evolutionary algorithms. Two types of algorithms (depth evolution and breadth evolution) are used to enhance the complexity and scope of basic instructions, generating more complex and diverse high-quality instruction data by ChatGPT.

LIMA \cite{zhou2023lima} is a dataset consisting of 1,000 training samples, 300 test samples, and 50 development samples. In addition to the manually authored samples, samples collected from Q\&A websites are strictly selected by human. Despite its modest scale, LIMA stands out for its meticulous compilation and design. LLMs fine-tuned on LIMA show remarkable ability in following instructions and adapting to unseen tasks.

DollyV2 \cite{DollyV2} is a dataset of 15,000 human-authored instruction-response pairs covering tasks such as brainstorming, classification, question answering, and summarization. The data was created by employees using only Wikipedia as a source, with restrictions against using web content or generative AI tools when crafting responses.

P3 \cite{P3} is a large-scale prompt collection that integrates 170 NLP datasets with 2,052 handcrafted prompts, enabling unified training and evaluation across diverse language tasks. These prompts, also known as task templates, convert conventional NLP tasks like question answering or text classification into natural language input-output pairs. The P3 dataset itself is formed by randomly selecting prompts from PromptSource and organizing data into triplets of inputs, answer choices, and targets.

\section{Data Selection Methods}
\label{selectors}
Formally, define an instruction dataset $X$ with a size of $n$, where $X = \{x_1, x_2, \dots, x_n\}$, and each $x_i$ represents an instance of instruction fine-tuning data. To employ a specific instruction data selection method $\pi$ from $X$ and choose a subset $S_{\pi}^{(m)}$ with a size of $m$ , we then use a predefined evaluation metric $Q$ to assess the quality of $S_{\pi}^{(m)}$. The obtained subset quality, as measured by the evaluation metric, can gauge the effectiveness of the chosen instruction data selection method. The process of designing the selection method can be seen as:
\begin{equation}
\label{eq1}
S^{(m)}_{\pi} = \pi \left( X \right)
\end{equation}
\begin{equation}
\label{eq2}
\pi^* = \arg\max_{\pi} Q\left(S^{(m)}_{\pi}\right)
\end{equation}

The classification of instruction data selection methods is based on what scoring rules the method uses and what model base it employs. The methods can be divided into the following four categories: methods based on system of indicators, trainable LLMs, powerful LLMs like ChatGPT and small models.

\subsection{Methods Based on a System of Indicators}
\label{indicators}

Methods based on a system of indicators aim to define a comprehensive set of evaluation metrics \( I_1, I_2, \dots, I_n \), each governed by a specific computational formula. These metrics are designed to assess different aspects of data quality or relevance. In certain cases, the metrics may incorporate deep learning models for feature extraction, which function as learned indicators.

For a given data instance \( x_j \), each metric \( I_i \) produces a score \( \text{score}_{ij} = I_i(x_j) \). These individual metric scores are then aggregated using a combination function \( G(\cdot) \), resulting in an overall score:

\begin{equation}
\label{eq3}
\text{score}_{j} = G \left( I_1(x_j), I_2(x_j), \dots, I_n(x_j) \right)
\end{equation}

Once the scoring function \( G \) is defined, the system can be applied to the entire dataset \( X \) to evaluate each data instance. Based on a predefined threshold \( \tau \), a high-quality subset \( S_{\pi} \) can be selected as follows:

\begin{equation}
\label{eq4}
S_{\pi} = \{x \in X \mid G(x) > \tau\}
\end{equation}

This indicator-based framework provides a structured and scalable approach to data selection, enabling the integration of both heuristic and learned metrics.

INSTRUCTMINING \cite{InstructionMining} is a linear rule-based method for assessing the quality of instructional data. This approach initially identifies key natural language metrics, such as instruction length, perplexity, reward scores, KNN-i\cite{Sentence-BERT,knn}, among others. These metrics are subsequently utilized to develop a linear equation. To explore the correlation between data quality and these metrics, and to define the parameters of equation, comprehensive fine-tuning experiments are conducted. Diverse quality datasets are partitioned into subsets, combined, and then employed in the fine-tuning of large-scale models. Quality labels for each subset are derived by evaluating performance of the model on a test set. The least squares method is applied to these experimental outcomes to estimate the parameters in INSTRUCTMINING. This involves fitting a linear equation to the loss observed when evaluating the model against the test set.Once the parameters have been determined, this formula can be utilized to calculate the quality of instructions, thereby facilitating data selection.

InstructionGPT-4 \cite{InstructionGPT-4} is a data selection method designed for fine-tuning multimodal large language models. It surpasses MiniGPT-4 \cite{MiniGPT-4} in various assessments using less data. The first step, metrics like the CLIP Score\cite{CLIP}, instruction length, and others are used. Visual and textual data are encoded into vectors and then subjected to dimensional reduction, which is regarded as special metrics. These metrics are combined into a vector $e$. The vector is then fed to a trainable data selector, such as a multi-layer perceptron or a self-attention network. This approach, similar to the method \cite{InstructionMining} of calculating quality labels, employs clustering algorithms to segment datasets. Quality labels are assigned post fine-tuning and evaluation of each subset.

DQ \cite{DatasetQuantization} is an innovative data compression approach originally developed for large-scale computer vision datasets, and later adapted for use in the LLM domain. The approach involves several key steps: Initially, a gain function $P(x)$ is defined as: 

\begin{equation}
\label{eq6}
P(x) = \sum_{p \in S} ||f(p) - f(x) ||^2_2 - \sum_{p \in D \setminus S}||f(p) - f(x)||^2_2 ,
\end{equation}
incorporating a feature function $f(.)$ — analogous to a metric — and using the current subset $S$ and the entire dataset $D$. This gain function essentially forms a metric set function $G(.)$. The dataset is then iteratively partitioned into non-overlapping subsets, guided by the gain function to maximize the defined gain. Subsequently, a representative sample from each subset is uniformly selected, ensuring coverage of the entire dataset while optimizing for maximum data diversity. This approach prioritizes maintaining the overall diversity of the dataset.

\subsection{Methods Based on Trainable LLMs}
\label{Trainable}

This section introduces methods that leverage trainable large language models (LLMs) to develop scoring functions for instruction data selection. In these methods, the LLM itself is fine-tuned or adapted to act as a data selector, assigning a score to each instruction instance:

\begin{equation}
\label{eq7}
\text{score}_{i} = \text{LLM}_{\text{trainable}}(x_i)
\end{equation}

Unlike indicator-based methods, which rely on predefined or hand-crafted metrics, trainable LLM-based methods learn to evaluate instruction quality directly from data. These methods not only focus on scoring individual instructions but also emphasize the alignment between the data selection process and the downstream performance of the fine-tuned model. Specific methods and frameworks that implement this paradigm are discussed in the following sections.

Instruction following difficulty (IFD) \cite{IFD} represents a trainable data selection approach grounded in the hypothesis that large language models (LLMs) can first be trained on a carefully curated subset of instruction data to acquire basic instruction-following capabilities. This foundational skill then enables the model to better estimate instruction difficulty and overall data quality. 

The method begins by fine-tuning an LLM on a small, clustered subset of the instruction dataset. It then introduces a novel metric, also named IFD, which quantifies the challenge of responding to a given instruction. Specifically, it compares the model's ability to generate the correct answer both with and without the given instruction. The \emph{Conditioned Answer Score} is computed as:

\begin{equation}
\label{eq8}
s_{\theta}(A|Q) = \frac{1}{N} \sum_{i=1}^{n} \log P(w_i \mid Q , w_1,\dots,w_{i-1}),
\end{equation}

where $Q$ is the instruction and $A$ is the target answer consisting of tokens $w_1, \dots, w_n$. This measures the model's likelihood of generating $A$ when conditioned on $Q$.

The final IFD score is defined as the ratio between the conditioned and unconditioned scores:

\begin{equation}
\label{eq9}
r_{\theta}(Q,A) = \frac{s_{\theta}(A|Q)}{s_{\theta}(A)},
\end{equation}

where $s_{\theta}(A)$ is the score of generating $A$ without conditioning on $Q$. Higher IFD scores indicate samples where the instruction significantly influences the output, thereby reflecting greater instruction-following difficulty. By applying a threshold on the IFD score, the method selects informative instructions to further refine the model in subsequent fine-tuning stages.

Instruction Backtranslation \cite{Instruction_Backtranslation} is a method for generating and filtering instructions. Starting with a baseline instruction-following model and a web corpus, the model generates instructions for each web document, forming a dataset. Then the model is fine-tuned with seed instructions to gain basic capabilities. And it  autonomously scores each instruction, with those exceeding a set threshold forming a high-score subset for further fine-tuning. This iterative process enhances instruction generation and filtering efficiency.

Nuggets \cite{One_Shot_Learning} is a framework that implements a dual-phase approach. Initially, it employs a variety of predefined tasks to evaluate the proficiency of a LLM across multiple scenarios, a process known as zero-shot scoring. Subsequently, each entry in the instruction dataset is utilized as a unique prompt in a one-shot manner. These prompts are presented before the predefined tasks, and the performance of LLM is reassessed, a step termed one-shot scoring. This approach harnesses the disparity between one-shot and zero-shot scores to calculate a definitive 'gold score' for each instruction. Upon acquiring gold scores for all instructions, those constituting the highest-scoring subset are chosen as the 'gold subset'. This subset is then directly employed in the fine-tuning of the model. This methodology exploits the inherent contextual learning capabilities of extensive models.

DIVERSEEVOL \cite{Self-Evolved} is an innovative iterative data selection mechanism. It leverages large-scale models like LLaMa to generate embedding vectors for instructional data. The mechanism employs the k-center-greedy algorithm\cite{k-center-greedy} to facilitate diversity in selecting a subset of data for fine-tuning the LLaMa model. This procedure is repetitively applied, progressively enlarging the chosen subset, culminating in the creation of a high-quality instructional dataset.

TEGIT \cite{TeGit} is a novel method for generating high-quality instruction fine-tuning data. Particularly notable is their method of filtering instruction data. Utilizing ChatGPT, a small corpus of documentation is transformed into a format suitable for instruction data, forming a meta-dataset. This dataset then serves to train two Llama2 models - one functioning as a task generator and the other as a task discriminator. The role of generator is to devise tasks from provided texts, while the discriminator evaluates these tasks, ensuring their quality.

Active Instruction Tuning \cite{ActiveIT} is a method that focuses on task sensitivity selection to enhance large model fine-tuning with fewer tasks while improving out-of-task generalization. This method introduces the concept of Prompt Uncertainty, which is determined by generating k perturbed instructions through the random deletion of words from an original instruction. The output of LLM probability deviation is then averaged for these k perturbed instructions. Tasks that exhibit higher Prompt Uncertainty are prioritized for instruction fine-tuning, using the extent of Prompt Uncertainty as a measure of task uncertainty.

\subsection{Methods Based on Powerful LLMs like ChatGPT }
\label{Powerful}

This section introduces the methods of using powerful LLMs such as GPT-4 and ChatGPT as data selectors. The approachs primarily involve designing prompt templates and leveraging the capabilities of LLMs to assess the quality of instruction data.

\begin{equation}
\label{eq10}
S_{\pi} = \{x | ChatGPT(score | prompt , x) , score > \tau\}
\end{equation}

ALPAGASUS \cite{AlpaGasus} is an innovative data filtering methodology aimed at improving efficiency and accuracy in Instruction-Following Task (IFT) data curation. This approach utilizes a well-designed prompt applied to ChatGPT, to assess the quality of each data tuple, comprising instruction, input, and response. The methodology focuses on excluding tuples that fall below a predefined quality threshold. When this filtering process was applied to a substantial dataset, it was observed that a considerable portion of the data was compromised by quality issues. Notably, the application of the LLM-based filtering process led to the development of a model that surpassed the performance of the original model, which was trained using the unfiltered dataset with instruction-based fine-tuning.

INSTAG \cite{InsTag} is an automated instruction tagging method that employs ChatGPT to generate detailed, open-ended labels for instructions, enhancing the diversity and complexity of the resulting subset. The process involves annotating data with labels that reflect the semantic and intent of each instruction and normalizing them for a label-based selection approach. This approach follows a complexity-first diversification sampling strategy. It starts by sorting queries in descending order of label count and then iteratively adds queries to a subset based on the uniqueness of their labels, until the subset reaches a desired size N. The result is a curated set of queries with a higher number of labels, indicating greater complexity and diversity.

To enhance the distribution and quality of datasets, \cite{LIFT} propose the method LIFT to reduce the redundancy of samples. It involves two stages: expanding dataset distribution and curating the diversity and quality of the dataset. Initially, ChatGPT enhances the data by generating diverse instructions and vectorizing them. A subset is then selected based on row variances. Secondly, ChatGPT scores instructions for accuracy, interpretability, clarity, difficulty, and length. The initial subset is re-selected based on these scores.

DEITA \cite{DEITA} integrates a multifaceted approach for selecting instruction data, focusing on complexity, quality, and diversity. Complexity describes factors such as the length, difficulty, and complexity of the instruction. And quality  captures the accuracy of the output. Utilizing the WizardLM technique, ChatGPT is employed to augment instructions, which are then evaluated for complexity and quality. These evaluations involve scoring instructions based on complexity, using a specially trained complexity scorer, and assessing the output quality. Each instruction in the dataset is assigned complexity scores ($c$) and quality scores ($q$), and a comprehensive score is calculated by multiplying these two metrics. The dataset is then organized according to these comprehensive scores and vectorized for further analysis. To ensure diversity, a subset is created by adding samples that exceed a set distance threshold ($\tau$) from their nearest neighbor in the subset. This process continues until the subset reaches a predetermined size.

InstructionNode \cite{InstructionNode} is a tree-instruct method, which improves the  quality of instruction by enhance the complexity of it. DEITA \cite{DEITA} uses this method to measure instruction complexity and sets thresholds for instruction filtering. Tree-instruct employs GPT-4 to generate semantic parse trees for instruction data, using the number of nodes in the tree as a measure of complexity. The complexity is enhanced by adding nodes to the tree, and then GPT-4 is used to convert the new tree back into natural language, resulting in new high-quality instructions.

WaveCoder \cite{WaveCoder} is a code-focused LLM enhanced by an instruction improvement technique. Its training incorporates generated data, with the data filtering phase being particularly significant. Following data generation, a LLM-based discriminator, leveraging GPT-4, evaluates instruction data against established criteria divided into subtopics. This approach enables finer control over the filtering process, effectively eliminating low-quality instruction instances.

\subsection{Methods Based on Small Models }
\label{Reward}

This section introduces methods involving the use of external small models as scorers, or the transformation of instructions into embedding vectors using small models followed by further processing. Generally, these methods are quite comprehensive. The scoring filtering or embedding generation by Small Models often serves only as a part of the entire methodological process.

MoDS \cite{MoDS} focuses on instruction selection through three criteria: quality (instructional data fidelity), coverage (variety of instruction types), and necessity ( impact of instruction on LLM fine-tuning). The process unfolds in four key steps. Firstly, employing a reward model to evaluate the instruction dataset quality, selecting a subset denoted as $D_{h}$, comprising instructions that surpass a predefined quality threshold. Secondly, utilizing the k-center-greedy algorithm \cite{k-center-greedy} to identify seed instructions, which ensures a diverse and representative sample of the instruction dataset. Thirdly, fine-tuning a pre-trained LLM with the seed instructions. Subsequently, this refined model is applied to $D_{h}$ to generate a new dataset, $D_{inference}$. This dataset is then evaluated using the reward model, which identifies instructions critical for learning of the LLM by focusing on those with lower scores. A threshold is established to select Augmented Instruction Data, which is specifically tailored to enhance performance of the model.Finally, the seed instructions with the Augmented Instruction Data to form a high-quality instruction subset that is poised to facilitate effective fine-tuning of the LLM.

The work \cite{0.5data} proposes a coreset-based and task-related data selection method: firstly, obtain sentence embeddings of samples by pre-trained language models (like BERT), then select the center point by applied unsupervised clustering on these embeddings, and finally use the KCenterGreedy \cite{KCentergreedy} algorithm to retrieve core samples from the given dataset. This method effectively reduces the required training data volume, maintaining or potentially enhancing model performance.

\subsection{Comparative Analysis of Data Selection Methods}
\label{comparison}
In this subsection, we provide an in-depth comparison of the existing data selection methods across several key dimensions: model dependence, scalability, domain adaptability, interpretability, and selection effectiveness. The comparison results are shown in Table~\ref{tab:method_comparison}.

\begin{table}[ht]
\centering
\caption{Comparison of Instruction Data Selection Methods}
\label{tab:method_comparison}
\begin{tabular}{ccccc}
\toprule
Dimensions & Methods in Sec.\ref{indicators} & Methods in Sec.\ref{Trainable} & Methods in Sec.\ref{Powerful} & Methods in Sec.\ref{Reward} \\
\midrule
\makecell{Model \\ Dependence} & \makecell{Model-agnostic \\ and heuristic} & \makecell{Requires supervised \\ model training} & \makecell{Depends on external \\ black-box APIs \\ (e.g., GPT-4)} & \makecell{Uses small pre- \\ trained models \\ (e.g., BERT, TinyLM)} \\
\midrule
Scalability & High & Low & Low & Moderate \\
\midrule
\makecell{Domain \\ Adaptability} & Low & High & Medium & Low to Medium \\
\midrule
Interpretability & High & Low & Low & Moderate \\
\midrule
\makecell{Selection \\ Effectiveness} & Variable  & High  & High & Medium \\
% GPU usage & Zero & Huge (e.g., A100) & Zero &  Medium (e.g., 4090) \\
\bottomrule
\end{tabular}
\end{table}

The first dimension, model dependence, captures how much each method relies on external or trainable models. Indicator-based approaches are generally model-agnostic, relying on heuristic or statistical features derived directly from the data, without needing to query or train large models. In contrast, trainable LLM-based methods require supervised training of a dedicated selector or reward model, often built on large-scale LLMs, making them heavily model-dependent. Methods that use powerful LLMs like GPT-4 operate in a black-box manner, scoring or filtering instructions via external APIs, which introduces reliance on proprietary infrastructure. Small model approaches strike a balance by leveraging small-scale pretrained models (e.g., BERT or TinyLM), reducing resource demands while maintaining some model-based reasoning.

Indicator-based methods clearly excel in scalability: they are fast, parallelizable, and suitable for large-scale filtering without model inference. Trainable methods, by contrast, face serious scalability challenges due to the high computational cost of model training and inference. Black-box LLM-based methods suffer from scalability bottlenecks due to API latency, usage limits, and costs, making them impractical for processing massive datasets unless sufficient resources are available. Small model methods offer a middle ground—they are not as scalable as pure heuristics but can be applied to larger datasets with reasonable efficiency.

Domain adaptability varies notably among these methods. Trainable methods exhibit the highest adaptability, as their learning process can directly align with new domains through supervised fine-tuning on relevant data. Powerful LLMs also adapt reasonably well in few-shot or zero-shot scenarios, depending on the prompt design and instruction diversity. Indicator-based methods, however, tend to be less adaptive since their features are usually handcrafted and not easily transferable across domains. Small models may offer limited adaptability depending on their pretraining corpus and whether further domain-specific fine-tuning is conducted.

% When it comes to interpretability, heuristic indicator methods clearly lead, offering intuitive, transparent criteria for data selection (e.g., length, diversity, perplexity). This is important for tasks requiring human oversight or explainable decision-making. In contrast, both trainable and black-box LLM-based methods operate as opaque systems, making it difficult to trace why certain instances are selected. Small models offer moderate interpretability—they can be analyzed via model outputs and saliency techniques but are not as transparent as heuristics.

When it comes to interpretability, heuristic indicator methods clearly lead, offering intuitive, transparent criteria for data selection (e.g., length, diversity, perplexity). This is important for tasks requiring human oversight or explainable decisions. In contrast, both trainable and black-box LLM-based methods operate as opaque systems, making it difficult to trace why certain instances are chosen. Small models offer moderate interpretability—they can be analyzed via model outputs and saliency techniques but are less transparent than heuristics.

Finally, the selection effectiveness dimension compares how well each method identifies high-quality instruction data. Both trainable and powerful LLM-based methods tend to produce strong results, as they explicitly optimize or evaluate instruction usefulness using powerful language models. Small models offer moderate performance—better than pure heuristics but generally not on par with full-scale LLMs. Indicator-based methods exhibit variable performance, with effectiveness largely dependent on the specific metrics used and the task domain.

In summary, no single method dominates across all dimensions. Practitioners must balance trade-offs among efficiency, generalizability, and interpretability when choosing a data selection strategy, or consider hybrid designs that combine the strengths of multiple approaches.

\section{Evaluation methods and Result Analysis}
\label{Evaluation}

To measure the effectiveness of different instruction data selection methods, several evaluation metrics are proposed and can be divided into three categories: winning rate, inner comparison, and external comparison. In this section, we first introduce each distinct evaluation and then compare different data selection methods across multiple benchmarks under these evaluations to analyze which methods achieve better performance.

\begin{table*}[!ht]
  \centering
  \caption{Performances of different selection methods on wining rate.}
  \begin{adjustbox}{max width=\textwidth}
    \begin{tabular}{ccccccccc}
    \toprule
    \multirow{2}[4]{*}{Selection methods} & \multicolumn{1}{c}{\multirow{2}[4]{*}{Model pairs for comparison}} & \multirow{2}[4]{*}{Training set} & \multicolumn{6}{c}{Win rate across various benchmarks} \\
\cmidrule{4-9}          &       &       & \multicolumn{5}{c}{Vicuna/Koala/WizardLM/Self-inst/LIMA} & \multicolumn{1}{l}{total WS} \\
    \midrule
    \multirow{5}[0]{*}{\makecell[c]{IFD \cite{IFD}}} & \multicolumn{1}{c}{Llama-7B(5\%), Llama-7B(full) } & \multirow{3}[0]{*}{alpaca} & \multicolumn{5}{c}{1.125/0.97/1.077/1/1.1} & 1.04 \\
    & \multicolumn{1}{c}{Llama-7B(10\%), Llama-7B(full) } &  & \multicolumn{5}{c}{1.037/1.055/1.114/1.123/1.103} & 1.097 \\
    & \multicolumn{1}{c}{Llama-7B(15\%), Llama-7B(full) } &  & \multicolumn{5}{c}{1/1.038/1.114/1.027/1.09} & 1.064 \\
    \cline{2-5}
          & \multicolumn{1}{c}{Llama-7B(10\%), Llama-7B(full)} & WizardLM & \multicolumn{5}{c}{1.1625/1.1278/1.1147/1.0278/1.1067} & 1.0971 \\
          & \multicolumn{1}{c}{Llama2-7B(5\%), Llama2-7B(full) } & alpaca & \multicolumn{5}{c}{1.5875/1.4889/1.4266/1.2937/1.4733} & 1.4311 \\
    \midrule
    Random Sampling & \multicolumn{1}{c}{Llama-7B(5\%), Llama-7B(full) } & alpaca & \multicolumn{5}{c}{-}                  & 0.9 \\
    \midrule
    \makecell[c]{InstructionGPT4 \cite{InstructionGPT-4}} & \multicolumn{1}{c}{miniGPT4(6\%), miniGPT4(full)} & cc\_sbu\_align & \multicolumn{5}{c}{-}                  & 1.167 \\
    \midrule
    \multirow{4}[0]{*}{\makecell[c]{Alpagsaus \cite{AlpaGasus}}} & \multicolumn{1}{c}{Llama-7B(9k), Llama-7B(full)} & \multirow{4}[0]{*}{alpaca} & \multicolumn{5}{c}{1.2125/1.0222/1.0596/1.0556/-} & 1.0658 \\
    & \multicolumn{1}{c}{Llama-7B(9k), Llama-7B(3k)} &       & \multicolumn{5}{c}{1.1/1.183/1.082/1.17/-} & 1.074 \\
     & \multicolumn{1}{c}{Llama-7B(9k), Llama-7B(6k)} &       & \multicolumn{5}{c}{1.05/1.072/1.05/1.087/-} & 1.082 \\
    & \multicolumn{1}{c}{Llama-13B(9k), Llama-13B(full)} &       & \multicolumn{5}{c}{1.2125/1.0167/1.133/1.0198/-} & 1.074 \\
    \midrule
    \makecell[c]{MoDS \cite{MoDS}}  & Llama2-7B(2k), Llama2-7B(full) & alpaca & \multicolumn{5}{c}{1.7125/1.5111/1.4725/1.369/1.4933} & 1.4786 \\
    \midrule
    \makecell[c]{InstructionMining \cite{InstructionMining}} & Llama-7B(2k), Llama-7B(full)  & dolly & \multicolumn{5}{c}{-}                  & 1.088 \\
    \bottomrule
    \end{tabular}%
    \end{adjustbox}
    % }
    \begin{tablenotes}
        \scriptsize
        \item '-' denotes values not reported in the original paper, and model(x) denotes tuning the model on the given training set using x samples or x\% of samples.
    \end{tablenotes}
  \label{win-rate}%
\end{table*}%

\subsection{Wining Rate}
\label{Rate}
To evaluate the effectiveness of the dataset selection methods, \cite{AlpaGasus} define the winning rate of the LLM fine-tuned with less samples (i.e., LLM-sub) compared to the LLM fine-tuned with full samples (i.e., base LLM) and is calculated as: 

\begin{equation}
\label{Wining-Rate}
\text{Win Rate}= \frac{Num(win) - Num(lose)}{Num(all)}  + 1 , 
\end{equation}
where Num(win) represents the number of wining cases, Num(lose) represents the number of losing cases, Num(all) represents the number of all cases in the testing benchmarks, and 1 indicates equal performance, values above 1 indicate improved performance, and values below 1 suggest degraded performance.

LLM-sub represents the LLM fine-tuned on the subset that filtered from the training set by the selection methods while the base LLM often involves 2 types: i) fine-tuned on the full training set and ii) fine-tuned on the same-scale subset filtered by the regular selections (e.g. random sampling and instruction length). The outputs of LLM-sub and the base LLM are rated by the judge on a scale from 1 to 10, and usually GPT4 is employed as the judge. To address the positional bias of the judge, IFD \cite{IFD} send the outputs of these two LLMs to the judge twice with different orders. According to this work, the wining cases refer to LLM-sub outperforming the base LLM in both times or winning in one and tying in the other. The losing cases refer to LLM-sub lagging behind the base LLM in both times or tying in one and losing in the other. The wining rates of different selection methods on the testing benchmarks are summarized in Table \ref{win-rate}.

\begin{table*}[!ht]
  \centering
  \caption{Inner comparisons of LLM tuned on subsets with itself tuned on full sets.}
   \begin{adjustbox}{max width=\textwidth}
    \begin{tabular}{cccc}
    \toprule
   Selection methods & Training set(samples) & Base model & \colorbox{cyan!8}{Win rate across various benchmarks} \\
\midrule
    &       &       & \colorbox{cyan!8}{total WS} \\
    \multirow{2}[2]{*}{\makecell[c]{activeIT \cite{ActiveIT}}} & selfinstruct(2k) & \multirow{2}[2]{*}{Llama-7B} & 1.107 \\
    & selfinstruct(full) &       & 1.293 \\
    \hline
    &       &       & \colorbox{cyan!8}{BBH/DROP/MMLU/Human-Eval/Avg} \\
    \multirow{2}[2]{*}{\makecell[c]{DQ \cite{DatasetQuantization}}} & alpaca(20\%)  & \multirow{2}[2]{*}{Llama-7B} & 32.7/26.7/39.8/9.2/27.1 \\
    & alpaca(full) &       & 32.9/26.3/41.6/10/27.7 \\
    \hline
    &       &       & \colorbox{cyan!8}{Vicuna RS/Vicuna WTR/Koala RS/ Koala WTR} \\
    \multirow{4}[2]{*}{\makecell[c]{DIVERSEEVOL \\ \cite{Self-Evolved}}} & Dolly(1k) & \multirow{4}[2]{*}{Llama-7B} & 79.69/20/62.29/6.67 \\
          & Dolly(full) &       & 73.84/5/57.9/3.33 \\
          & SelfIns(1k) &       & 79.16/7.5/66.95/6.11 \\
          & SelfIns(full) &       & 73.03/2.5/69.5/3.89 \\
    \hline
     &       &       & \colorbox{cyan!8}{HellaSwag/ARC/TruthfulQA/MMLU} \\
    \multirow{2}[2]{*}{\makecell[c]{LIFT \cite{LIFT}}} 
& Platypus(15k random) &       & 0.82/0.607/0.438/0.625 \\
& Platypus(15k) &       & 0.844/0.643/0.49/0.645 \\
    \hline
    &       &       & \colorbox{cyan!8}{RTE/CB/ANLI R1/ANLI R2/ANLI R3} \\
    \multirow{2}[2]{*}{\makecell[c]{coreset \cite{0.5data}}} & P3(0.5) & \multirow{2}[2]{*}{Galactica-1.3B} & 74.73/73.21/49.6/41.9/43.75 \\
    & P3(full) &  & 76.17/75/44/35.7/39.42 \\
    \bottomrule
    \end{tabular}%
     \end{adjustbox}
    \begin{tablenotes}
        \scriptsize
        \item dataset(x) denotes tuning the base model on the x samples of the given set. 
    \end{tablenotes}
  \label{inner}%
\end{table*}%

\subsection{Inner Comparison}
\label{Inner_Comparison}
To evaluate the effectiveness of the dataset selection methods simply and directly, LLM-sub is compared with the same LLM, but it is fine-tuned on either the full training set or on a same-scale subset filtered by the regular selections. We refer to this evaluation method as inner comparison, because it only compares the LLM fine-tuned on a subset to itself. The inner comparison performances of different selection methods on the testing benchmarks are summarized in Table \ref{inner}.

Inner comparison reveals the trade-off between dataset size and quality. Although models fine-tuned on the full training set generally perform well, carefully selected subset methods can achieve comparable or even better performance with fewer samples, demonstrating improved efficiency. However, the performance gains vary across different methods and tasks, highlighting the importance of adjusting selection strategies based on the characteristics of downstream tasks and the diversity of instructions.

\subsection{External Comparison}
\label{External_Comparison}
Another simple and straightforward type of evaluation method is external comparison, which compares LLM-sub with the external LLMs (i.e. differ from the model of LLM-sub) on different testing benchmarks. The external comparison performances of different selection methods on the testing benchmarks are summarized in Table \ref{external}.

\begin{table*}[!ht]
  \centering
  \caption{External comparisons of LLM tuned on subsets with the other LLM.}
    \begin{tabular}{ccccc}
    \midrule
    Methods & Training set(samples) & Base model & MT-bench & AlpacaEval \\
    \midrule
    \multicolumn{1}{c}{gpt-4} &   \XSolidBrush  &  \XSolidBrush  & 8.99  & 95.28 \\
    \multicolumn{1}{c}{gpt-3.5-turbo} &   \XSolidBrush   &  \XSolidBrush   & 7.94  & 91.36 \\
    \multicolumn{1}{c}{alpaca-13B} & alpaca & Llama-13B & 4.53  & - \\
    \hline
    \multicolumn{1}{c}{NUGGETS\cite{One_Shot_Learning}} & alpaca(7.5k) & Llama-7B & 5.34  & - \\
    \hline
    TAGLM-13B-v1.0 \cite{InsTag} & \multicolumn{1}{c}{\multirow{6}[0]{*}{mixture(6k)}} & \multicolumn{1}{c}{Llama-13B} & \multicolumn{1}{c}{6.44±0.04} & 72.8 \\
    TAGLM-13B-v2.0 \cite{InsTag} &       & \multicolumn{1}{c}{Llama2-13B} & \multicolumn{1}{c}{6.55±0.02} & - \\
    \cline{3-5}
    \multicolumn{1}{c}{Instruction Length} &       & \multicolumn{1}{c}{\multirow{4}[0]{*}{Llama-13B}} & 5.89  & - \\
    \multicolumn{1}{c}{Random Sampling} &       &       & 5.84  &  -\\
    \multicolumn{1}{c}{IFD\cite{IFD}} &       &       & 5.91  & - \\
    \multicolumn{1}{c}{Instruction Node\cite{InstructionNode}} &       &       & 5.65  & - \\
    \hline
    \multicolumn{1}{c}{DEITA \cite{DEITA}} & mixture(10k) & Llama2-13B & 6.79  & 81.09 \\
    \bottomrule
    \end{tabular}%
    \begin{tablenotes}
        \scriptsize
        \item 'mixture' denotes a combined dataset of WizardLM(Alpaca), WizardLM(ShareGPT), UltraChat, and ShareGPT. 
    \end{tablenotes}
  \label{external}%
\end{table*}%

External comparison helps reveal the generalization ability of data selection methods across different base models and architectures. For example, as shown in Table \ref{external}, IFD\cite{IFD} demonstrates robust performance when transferring from Llama-7B to Llama2-7B, indicating that the selected data subsets contain general instruction information beneficial to multiple models. However, this evaluation also exposes some limitations: certain methods that perform well in inner comparison show diminished advantages in external comparison, possibly due to overfitting to the training distribution of specific models. For instance, although Alpagsaus \cite{AlpaGasus} performs excellently in inner comparison, its performance is unstable when compared against multiple external LLMs.

\subsection{Result Analysis}
\textbf{The proposed selection methods outperform the regular selections, which proves the significance of the data selection in instruction tuning.} As shown in Tabel \ref{win-rate} and \ref{inner}, TAGLM-13B-v1.0 and IFD outperform the regular selections on MT-benc based on instruction length and random sampling when tuning Llama-13B on the mixture dataset, and Alpagsaus and IFD outperform the random sampling on total WS when tuning Llama-7B on alpaca.

\textbf{The more advanced LLM (Llama2-7B) achieves higher performance than the standard LLM (Llama-7B)}, both of which are fine-tuned on the same subset. Table \ref{external} shows that TAGLM-13B-v2.0 outperforms TAGLM-13B-v1.0, and Table \ref{win-rate} shows that Llama2-7B(5\%) outperforms Llama-7B(5\%) when employing IFD selection method. These improvements are attributed to the inherent sophistication of the advanced LLMs, leading to higher learning efficiency on the subset.

\textbf{Instruction tuning the specific LLM on a larger subset does not necessarily guarantee performance improvement.} This may be related to the inherent features of the selection methods. As shown in Table \ref{win-rate}, the performance of Llama-7B does not improve with the increase of the subset size when applying the method IFD in the training set alpaca. However, the performance of Llama-7B improves with the increase of the subset size when applying the method Alpagsaus in the same training set.

The above findings indicate that no single method holds an absolute advantage across all tasks. Instead, hybrid approaches—such as using metrics for preliminary filtering followed by detailed evaluation by large language models (LLMs)—may represent a more promising direction. Future evaluations should further take into account the characteristics of downstream tasks and the types of instructions, rather than relying solely on coarse-grained metrics like overall accuracy or win rate.

\section{Discussion and Open Challenges}
\label{Conclusion}

This paper provides a comprehensive overview of methods and challenges in instruction data selection for LLM tuning, emphasizing the crucial role of high-quality data during the fine-tuning stage. We review existing instruction datasets and their construction approaches, highlighting common issues such as imbalanced data distribution and inconsistent quality. On this basis, we categorize four representative types of data selection methods. Indicator-based methods can effectively assess sample-level quality but often overlook dataset-level characteristics. Trainable LLM-based methods adapt to model-specific preferences through scoring functions. Methods based on powerful external LLMs (e.g., GPT-4) offer superior general scoring capacity. Lastly, small model-based methods often adopt modular pipelines to approximate high-cost selections efficiently.

We also introduce three evaluation paradigms: win rate (against reference data), inner comparison (vs. full/small subsets with the same model), and external comparison (vs. other LLMs). Although existing methods show promising results, several challenges remain:

\textbf{Lack of unified evaluation standards.} As detailed in Section \ref{Evaluation}, different methods adopt different evaluation schemes and benchmarks, making it difficult to compare across techniques fairly. Future research should aim to develop standardized, automatic evaluation protocols that comprehensively assess selection quality under consistent conditions.

\textbf{High computational cost and dependence on large models.} When filtering hundreds of thousands of instructions, methods using large LLMs or APIs (e.g., GPT-4) can be prohibitively expensive and slow. There is a pressing need for efficient selection approaches that reduce reliance on external models, for example, by leveraging distilled or lightweight selectors that mimic the behavior of larger models.

\textbf{Limited coverage across domains and languages.} Most current selection frameworks are designed for English and general-purpose tasks. Extending them to other languages (e.g., Chinese, Arabic) and specialized domains (e.g., biomedical, legal) remains an open challenge.

To concretize the future directions above, we propose several specific and actionable research paths:

First, addressing the \textbf{evaluation inconsistency}, future work should build a standardized benchmark suite for instruction selection. This includes diverse downstream tasks (reasoning, summarization, code generation), a variety of model architectures (e.g., LLaMA, Mistral), and multilingual datasets. In addition, we advocate for \textit{task-conditioned evaluation}, where selection quality is judged by its impact on real-world applications rather than just coarse-grained win rates.

Second, to improve \textbf{efficiency and scalability}, a promising strategy is to develop \textit{distillable selector models}—compact models trained to emulate powerful LLMs' selection behavior. Another approach is \textit{retrieval-augmented pruning}, in which fast heuristics narrow down candidate data, followed by LLM-based fine filtering only on top-ranked items, significantly reducing total cost.

Third, to enhance \textbf{domain and language adaptability}, future methods can integrate \textit{multilingual encoders} (e.g., XLM-R) and incorporate \textit{domain-specific priors}, such as metadata or instruction patterns. Modular design would allow swapping components to suit medical, legal, or programming domains, supporting cross-lingual and cross-domain generalization.

Lastly, we strongly encourage the \textbf{open-sourcing of reproducible selection pipelines}, including code, pretrained selector models, scoring results, and evaluation benchmarks. This will ensure fair comparisons and promote rapid iteration across the community.

\textbf{In summary}, instruction data selection is a critical but underexplored component in the instruction tuning pipeline. We hope this survey and analysis serve as a solid foundation to guide further research and practical improvements in making LLMs more instruction-following and efficient.

\begin{acks}
This work is supported by National Key R\&D Proram of China (No.2022ZD0160602 and No.2022YFF0902703), the National Natural Science Foundation of China (No.62472121), and the Special Funding Program of Shandong Taishan Scholars Project. 
\end{acks}

\printbibliography


\end{document}